\documentclass{am2013}

\usepackage{epsfig}
\usepackage{float}
\usepackage{amsmath}
\usepackage{setspace} 

\title{An alternative to SVM Method for Data Classification }

\author{Lakhdar Remaki\\
\vskip 2mm {\small
Departement of Mathematics and Computer Science, \\
Alfaisal University, KSA\\
lremaki@bcamath.org 
}
}

\begin{document}
 
\maketitle

\begin{abstract}

Support vector machine (SVM), is a popular kernel method for data classification that demonstrated its efficiency for a large range of practical applications. The method suffers, however, from some weaknesses including; time processing, risk of failure of the optimization process for high dimension cases, generalization to multi-classes, unbalanced classes, and dynamic classification. In this paper an alternative method is proposed having a similar performance, with a sensitive improvement of the aforementioned shortcomings. The new method is based on a minimum distance to optimal subspaces containing the mapped original classes.

\end{abstract}

\section*{keyword}
SVM, POD, kernels


\section{Introduction}
 
Machine learning (ML) is the science of algorithms that can automatically improve by learning from a set of data, see for instance \cite{Ethem20}. Classification problems are of a big importance in practice where \text{ML} techniques are extensively used.  Kernel methods are one of the most popular subset of data classifiers, and the Support Vector Machine (SVM) is one of the widely used technique of this family. SVM method is developed in the framework of statistical learning theory. The first work is back to Vapnik’s theory on the Structural Risk Minimization principle \cite{Vap95,Vap98,CorVap95}. The main principle of kernel methods, is to map the initial data, called attributes, into a higher dimension vector space called features space, where data are easily separated. The SVM technique is designed to separate linearly separable data, and when this is not the case, the mapping intends to make the data linearly separable in the feature space. The mapping function is not defined explicitly, which is a tedious task, but rather a kernel is used as long as the method steps can be expressed by a dot product. The existence of such kernels is guaranteed by the Mercer's theory \cite{Merc909,CorVap95} under some conditions. The SVM method is very popular for its demonstrated performance, however as any methods it has some shortcomings that can  be summarized as follow; it is an optimization problem, the process can then fail for high dimension even if the cost function is a quadratic (convex function). The generalization for multi-classes is time consuming since the commonly one-against-one combined to one-against-all used strategy requires calculating a total of $n + \frac{n(n-1)}{2}$ separating hyperplanes, $n$ being the number of classes. Imbalanced classes management is another issue, and finally a dynamic classification where new classes can form or disappear, all hyper-plans have to be re-calculated. As an alternative, this paper proposes a new kernel method based on a minimum distance to optimal subspaces (in the sense of smallest subspace containing the mapped classes). This method performs as well as the SVM, with the following advantages; The method doesn't require any optimization process, which makes it more robust, the complexity grows linearly with the number of classes, which reduces sensibly the processing time, no need to any recalculation for new classes in dynamic classification,  and the imbalanced classes is solved by decomposing the concerned subspace into sub-subspaces of a likewise dimensions. 
The subspaces in the features space are obtained using POD (proper orthogonal decomposition) and since distances to subspaces are calculated by projections they are expressed by dot products, and then can be calculated through the kernel.  

The paper is organized as follows, in chapter 1 an overview of the POD method is provided, in chapter 2 the new method is described, in chapter 4 2D tests are performed for validation and conclusions are drawn in chapter 5.

\section{Proper Orthogonal Decomposition (POD): Overview}
\label{POD-overview}
\noindent 
POD is a technique that aims to represent a big amount of data by a reduced number of basis elements built from the data. More precisely, if the data are stored in a $m\times n$ matrix $A$ with $n$ being a large number, is it possible to represent the $n$ columns of $A$ by a $p$ orthonormal vectors with $p$ very small compared to $n$. The mathematical formulation can be summarized as follows, let $A=\left[ V^{1},V^{2},...,V^{n}\right] $, the problem is to find $B=\left[ W^{1},W^{2},...,W^{p}\right] $ an orthonormal set of $p$ vectors that better represent the $n$ columns $A$ in the following sense,

 $W^{1}$ satisfying the optimization problem,
\begin{equation} \label{POD-opt}
W^{1} =Arg \bigg[Max_{W \in IR^{n}}\left( \sum_{i=1}^{n} \left(  \left\langle W, V_{i} \right\rangle \right)  ^{2}  \right)~~~ \textit{subject to} ~~~~ \Vert W \Vert^{2} = 1 \bigg].
\end{equation}

This constrained optimization problem can be solved using Lagrange multipliers:

\begin{equation} \label{Lagrange1}
L(W,\lambda) =  \sum_{i=1}^{n} \left(  \left\langle W, V_{i} \right\rangle \right)  ^{2} + \lambda(1-\Vert W \Vert^{2})~~~ (W,\lambda)\in IR^{m+1},
\end{equation}

and the solution is obtained by nullifying the gradient

\begin{equation} \label{grad}
\nabla L(W,\lambda) = 0 ~~~\text{in}~~~ IR^{n}\times IR. \nonumber
\end{equation}

Differentiating $L$ with respect of $W$

\begin{gather} \label{diff1}
\frac{\partial L}{\partial W_{j}} = \frac{\partial}{\partial W_{j}}\left( \sum_{i=1}^{n} \left( \sum_{k=1}^{m}  W_{k} V^{k}_{i} \right)  ^{2} + \lambda(1-\sum_{k=1}^{m}\left( W_{k} \right) ^{2}) \right) \nonumber \\
= \sum_{i=1}^{n}  \frac{\partial}{\partial W_{j}} \left( \sum_{k=1}^{m} W_{k} V^{k}_{i} \right)  ^{2} + \frac{\partial}{\partial W_{j}} \left( \lambda(1-\sum_{k=1}^{m}\left( W_{k} \right) ^{2}) \right) \nonumber \\
=\sum_{i=1}^{n}  2\left( \sum_{k=1}^{m} W_{k} V^{k}_{i} \right)V^{j}_{i} - 2 \lambda W_{j}  \nonumber \\
= 2\left( \sum_{k=1}^{n}  \left( \sum_{i=1}^{m} V^{k}_{i}V^{j}_{i} \right) W_{k} - \lambda W_{j}\right)   \nonumber \\
= 2\left( \sum_{k=1}^{n}  \left(AA^{T} \right)_{k,j} W_{k} - \lambda W_{j}\right)  \nonumber \\
\text{ $AA^{T}$ is symmetric we can permute the indexes} \\
= 2\left( \sum_{k=1}^{n}  \left(AA^{T} \right)_{j,k} W_{k} - \lambda W_{j}\right).  \nonumber 
\end{gather}

Then
\begin{gather} \label{diff1}
\frac{\partial L}{\partial W_{j}} = 0 ~~~\text{implies}~~~ \sum_{k=1}^{n}  \left(AA^{T} \right)_{j,k} W_{k} = \lambda W_{j}  \nonumber
\end{gather}

and then, by setting $W=W^{1}$ and $\lambda = \lambda_{1}$
\begin{gather} \label{diff1}
AA^{T}W^{1} =  \lambda_{1} W^{1}. \nonumber
\end{gather}
This shows that $W^{1}$ is the eigenvector of the matrix $AA^{T}$ and $\lambda_{1}$ is the associated eigenvalue. Now evaluate the maximum by substituting $W^{1}$,
\begin{gather}
\sum_{i=1}^{n}  \left\langle W^{1}, V_{i} \right\rangle ^{2}= \sum_{i=1}^{n}  \left( \sum_{k=1}^{m}W_{k}^{1} V_{k}^{i} \right)\left( \sum_{j=1}^{m}W_{j}^{1} V_{j}^{i} \right) \nonumber \\
= \sum_{i=1}^{n}  \left( \sum_{k=1}^{m} \sum_{j=1}^{m} W_{k}^{1} V_{k}^{i} W_{j}^{1} V_{j}^{i} \right) \nonumber \\
=  \sum_{k=1}^{m} \sum_{j=1}^{m} W_{k}^{1} W_{j}^{1}\sum_{i=1}^{n}  V_{k}^{i}V_{j}^{i}  \nonumber \\
=  \sum_{k=1}^{m} \sum_{j=1}^{m} W_{k}^{1} W_{j}^{1}(AA^{T})_{k,j} \nonumber \\
=\left\langle AA^{T}W^{1},W^{1} \right\rangle = \lambda_{1} \left\langle W^{1},W^{1} \right\rangle = \lambda_{1}.
\label{lambda1max}
\end{gather}

We seek $W^{2}$ to be the second best representative with $W^{2}$ perpendicular to $W^{1}$. We seek $W^{2}$ in the orthogonal space to the one spanned by $W^{1}$. This can be formulated as follows: 

\begin{equation} \label{POD-opt}
W^{2} = Arg \bigg[Max_{W \in IR^{n}}\left( \sum_{i=1}^{n} \vert \left\langle W, V_{i} \right\rangle \vert ^{2}  \right)~~~ \textit{subject to} ~~~~ \Vert W \Vert^{2} = 1 ~~\textit{and}~~ \left\langle W, W^{1} \right\rangle = 0 \bigg].
\end{equation}

Note the matrix $AA^{T}$ is symmetric, positive semi-definite, then it has $m$ eigenvalues $\lambda_{1}\geq \lambda_{2}\geq ... \lambda_{m} \geq 0  $ and $m$ corresponding eigenvalues $W^{1},...,W^{2}$ that can be chosen to be orthonormal, that is, 
\begin{gather}
\Vert W^{i} \Vert = 1 ~~~ \text{for} ~~~i=1,m  ~~~ \text{and} ~~~ \left\langle W^{i}, W^{j} \right\rangle = 0, ~~~\text{if}~~~ i\neq j. \nonumber 
\end{gather}
Since we look for a $W$ orthogonal to $W^{1}$, that is $W\in span\{W^{2},...,W^{m}\}$ it can be expended as
\begin{gather}
W= \sum _{k=2}^{k=m} \left\langle W,W^{k} \right\rangle W^{k}. \nonumber 
\end{gather}

Now estimate the quantity to maximize at an arbitrary $W$,
\begin{gather}
\sum_{i=1}^{n} \left\langle W, V^{i} \right\rangle ^{2} = \sum_{i=1}^{n} \left\langle \sum _{k=2}^{m} \left\langle W,W^{k} \right\rangle W^{k}, V^{i} \right\rangle ^{2} \nonumber \\
= \sum_{i=1}^{n} \left\langle \sum _{k=2}^{m} \left\langle W,W^{k} \right\rangle W^{k}, V^{i} \right\rangle \left\langle \sum _{k=2}^{m} \left\langle W,W^{j} \right\rangle W^{j}, V^{i} \right\rangle \nonumber \\
=\sum_{i=1}^{n} \left(  \sum _{k=2}^{m} \left\langle W,W^{k} \right\rangle \left\langle W^{k}, V^{i} \right\rangle  \right) \left(  \sum _{k=2}^{m} \left\langle W,W^{j} \right\rangle \left\langle W^{j}, V^{i} \right\rangle  \right) \nonumber \\
= \sum_{i=1}^{n} \sum _{k=2}^{m} \sum _{j=2}^{m} \left\langle W,W^{k} \right\rangle \left\langle W^{k}, V^{i}\right\rangle \left\langle W,W^{j} \right\rangle \left\langle W^{j}, V^{i}\right\rangle \nonumber \\
= \sum_{i=1}^{n} \sum _{k=2}^{m} \sum _{j=2}^{m} \left\langle W,W^{k} \right\rangle  \left\langle W,W^{j} \right\rangle \left\langle W^{j}, V^{i}\right\rangle \left\langle W^{k}, V^{i}\right\rangle \nonumber \\
= \sum _{k=2}^{m} \sum _{j=2}^{m} \left\langle W,W^{k} \right\rangle  \left\langle W,W^{j} \right\rangle \sum_{i=1}^{n} \left\langle W^{j}, V^{i}\right\rangle \left\langle W^{k}, V^{i}\right\rangle. \nonumber
\end{gather}
We can check by expending in terms of components that,
\begin{gather}
\sum_{i=1}^{n} \left\langle W^{j}, V^{i}\right\rangle \left\langle W^{k}, V^{i}\right\rangle = \left\langle AA^{T}W^{j},W^{k} \right\rangle, \nonumber
\end{gather}
and by orthogonality
\begin{gather}
\left\langle AA^{T}W^{j},W^{k}\right\rangle = 0~~~ \text{if}~~~ j\neq k,~~~ \text{and}~~~\left\langle AA^{T}W^{j},W^{j}\right\rangle = \lambda_{j},  \nonumber
\end{gather}
therefore
\begin{gather}
\sum_{i=1}^{n} \left\langle W, V^{i} \right\rangle ^{2} = \sum _{j=2}^{m} \left\langle W,W^{j} \right\rangle  \left\langle W,W^{j} \right\rangle \lambda_{j} \nonumber \\
=\sum _{j=2}^{m} \left\langle W,W^{j} \right\rangle ^{2} \lambda_{j} \nonumber \\
\leq \sum _{j=2}^{m} \left\langle W,W^{j} \right\rangle ^{2} \lambda_{2} \nonumber \\
= \lambda_{2} \sum _{j=2}^{m} \left\langle W,W^{j} \right\rangle ^{2} = \lambda_{2} \Vert W \Vert ^{2}= \lambda_{2}.
\end{gather}

Now if we substitute $W$ by $W^{2}$ following the same steps as in \eqref{lambda1max} we obtain
\begin{gather}
\sum_{i=1}^{n} \left\langle W^{2}, V^{i} \right\rangle ^{2} = \lambda_{2}, \nonumber
\end{gather}
which proves that the maximum is reached at $W^{2}$. We can repeat the same process and show that the $p$ vectors we looking for are the eigenvectors of $AA^{T}$. For more details see \cite{volkwein2013proper}.

\section{An Alternative Method to SVM}
\label{KOS-Method}
In this work a method based on kernel theory for data classification is proposed, the idea is to map the original data (attributes) into a bigger, in terms of dimension, features space as for the SVM, then optimal (in the smallest sense) subspaces that contain the mapped classes are built. finally, any unseen attribute vector will be classified by a simple projection onto the subspaces to determine the minimum distance and then the appropriate class. To build the subspaces the POD technique is used. Note that similar techniques like PCA (principle component analysis) are commonly used in kernel theory for data classification but as a pre-processing for features extraction and improvement \cite{Wang}. 

Suppose now that a suitable kernel $k(x,y)$ is given and let $\phi$ be the underlying mapping function. Note that selecting a kernel such that the output space $\textit{H}$ is of infinite dimension is recommended to increase the chance having the data mapped into distinct subspaces. Let $Y_{i}=\phi(X_{i})$ the mapped data. First form the correlation matrix 
\begin{equation}\label{kernel-ncent}
N(i,j)=Y_{i}Y_{j} =\phi(X_{i})\phi(X_{j}) = K(i,j).
\end{equation}
As we can see the POD correlation matrix corresponds to the mapping kernel.
The POD modes are then given by:

\begin{equation}\label{modes-ncent}
\psi_{i} = \sum_{k=1}^{N}V^{i}_{k}Y_{k},
\end{equation}
where $V_{i}$ is the $i^{th}$ eigenvector of the kernel matrix $N$

The decision criterion for an unseen element $\widehat{X}$ is based on the minimum distance of the mapped element $\widehat{Y}=\phi(\widehat{X})$ to the features subspaces  spanned by the POD modes. To calculate the distance, we need first to calculate the coordinate of $\widehat{Y}$ in the POD basis.
The coordinates $\alpha_{i},i=1,N$ are given by the projection on the POD modes 
\begin{equation}
\alpha_{i}=<\psi_{i},\widehat{Y}>
\end{equation}
But
\begin{equation}
\psi_{i} = \sum_{k=1}^{N}V^{i}_{k}Y_{k},
\end{equation}
Then
\begin{equation}\label{coord-ncent}
\alpha_{i}=<\sum_{k=1}^{N}V^{i}_{k}Y_{k},\widehat{Y}> = \sum_{k=1}^{N}V^{i}_{k}(Y_{k}\widehat{Y})=\sum_{k=1}^{N}V^{i}_{k}K(X_{k},\widehat{X})
\end{equation}

Using the Pethagorus rule the distance of $\overline{Y}$ to the POD subspace $F$ by

\begin{equation}\label{dist-ncent}
dis(\widehat{Y}, F) = \sqrt{ \Vert \widehat{Y}\Vert ^{2} - \sum_{i=1}^{N}\alpha_{i}^{2} }.
\end{equation}

With
\begin{equation}
\Vert \widehat{Y}\Vert ^{2} = \widehat{Y}\widehat{Y}= \phi(\widehat{X})\phi(\widehat{X}) = K(\widehat{X},\widehat{X}).
\end{equation}

Then
\begin{equation}\label{dist-cent}
dis(\widehat{Y}, F) = \sqrt{ \Vert \widehat{Y}-\overline{Y}\Vert ^{2} - \sum_{i=1}^{N}\alpha_{i}^{2} }.
\end{equation}

With
\begin{equation}
\Vert \widehat{Y}  - \overline{Y}\Vert ^{2}=  \widehat{Y}^{2} -2\widehat{Y}\overline{Y} + \overline{Y}^{2} = K(\widehat{X},\widehat{X})-\frac{2}{N}\sum_{j=1}^{N}K(\widehat{X},{X_{j}})+\frac{1}{N^{2}} \sum_{k=1}^{N}\sum_{j=1}^{N}K_{k,j}.
\end{equation}

\subsection{Imbalanced classes}
\label{imbapanced}
Imbalanced classes is one of the issues encountered in many classifiers, this can be naturally solved in the case of the proposed method. Using a reference size that can be the size of the smallest set in the original space, any class with a size at least double the reference size can be split into subsets of roughly equal sizes to the reference one. For the decision criterion, if the minimum distance is achieved with the subspace of one of the mapped subset, then the unseen element belong to the class.

\section{Validation}

\vspace{-0.19in}
The proposed method is verified and validated for the RBF kernel, first on 2D cases to see visually its capability for non-linear separation, and then two tests are performed and results are compared LIBSVM tool \cite{LIBSVM}.

\begin{figure}[H]
$$\begin{array}{ccc}
    \includegraphics[scale=0.30]{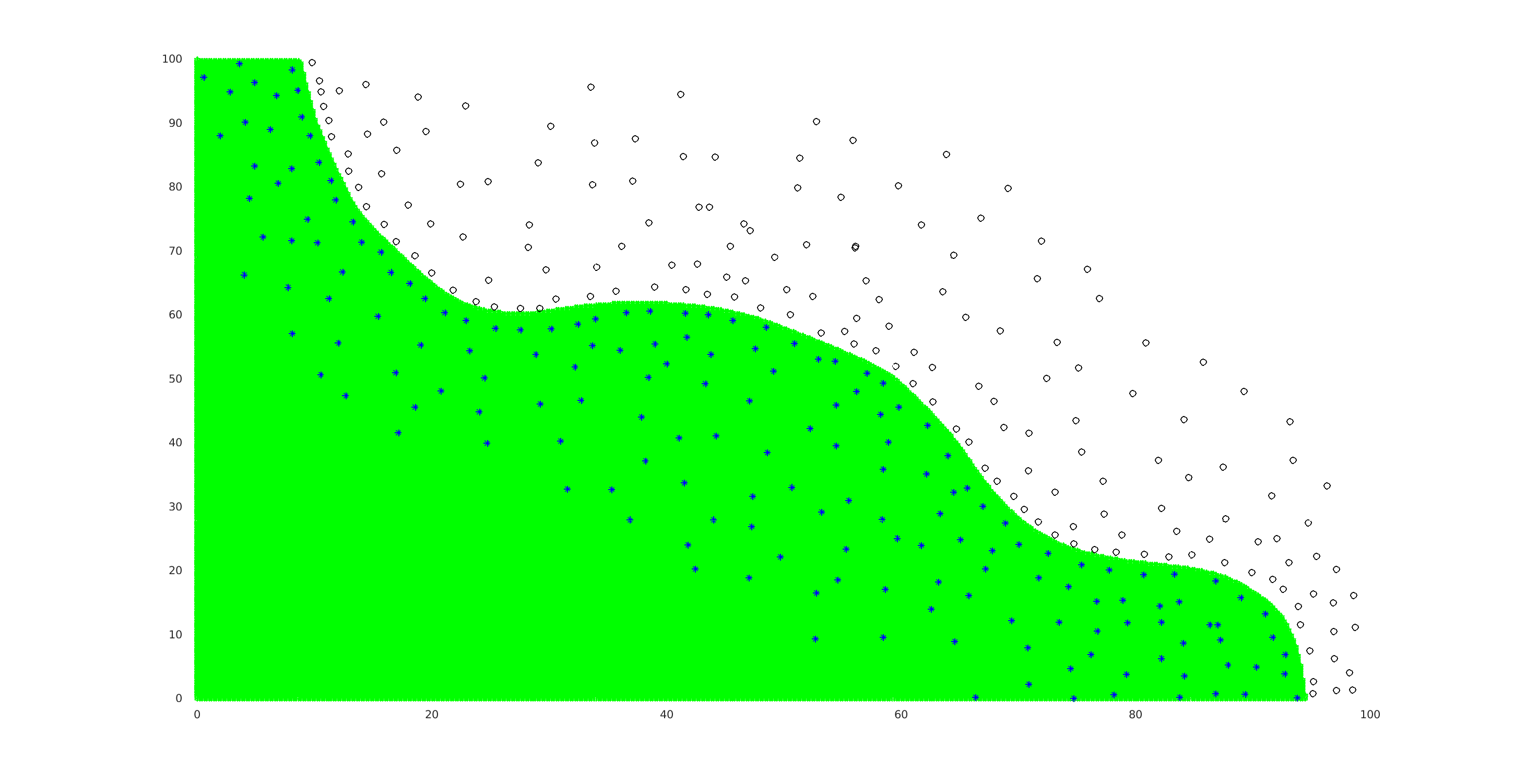} \\
\end{array}$$
\vskip-0.2cm
\caption{Connected sets case}
\label{connected}
\end{figure}

\begin{figure}[H]
$$\begin{array}{ccc}
     \includegraphics[scale=0.30]{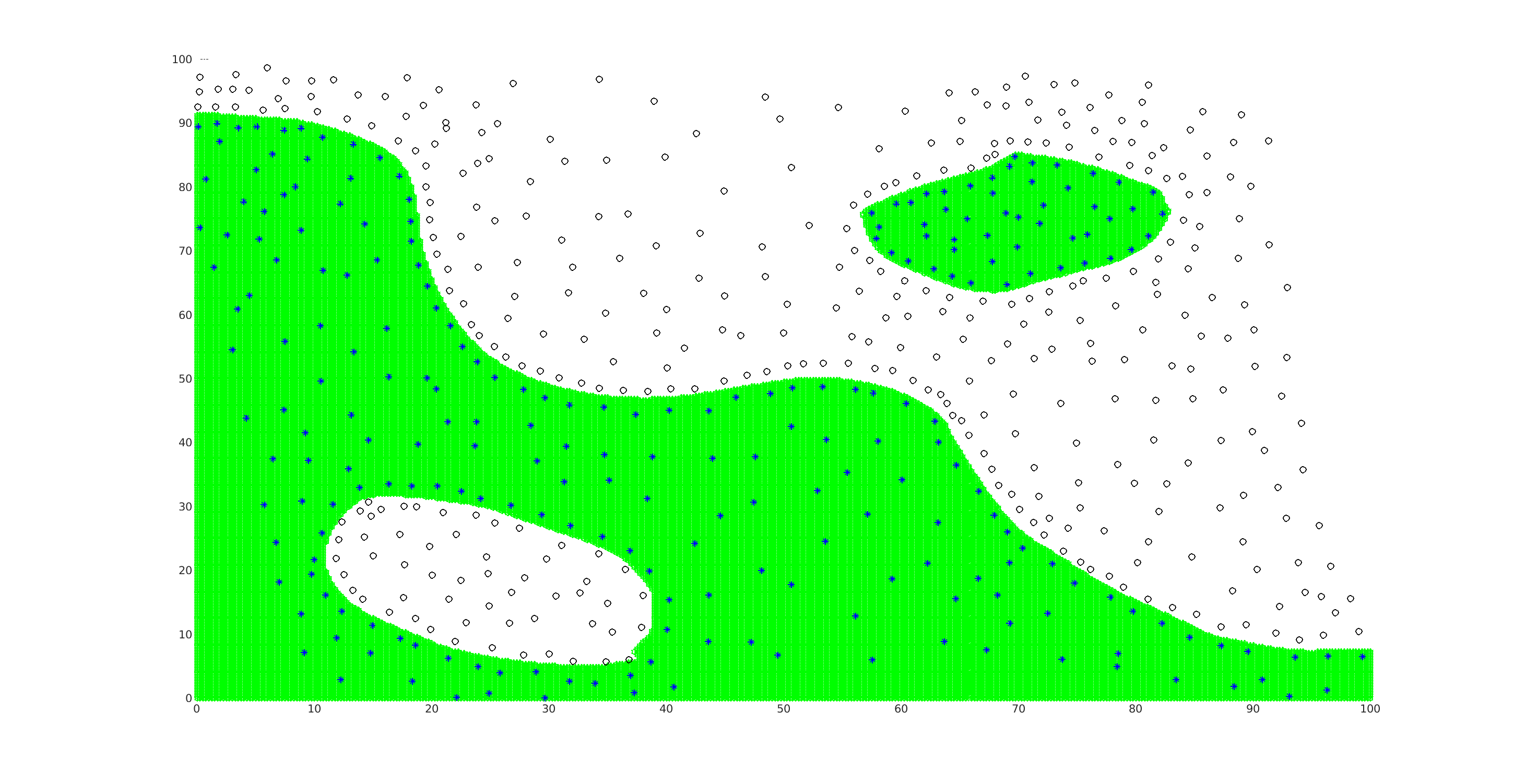}  \\
\end{array}$$
\vskip-0.2cm
\caption{Non-connected sets case}
\label{non-connected}
\end{figure}

\begin{figure}[H]
$$\begin{array}{ccc}
    \includegraphics[scale=0.30]{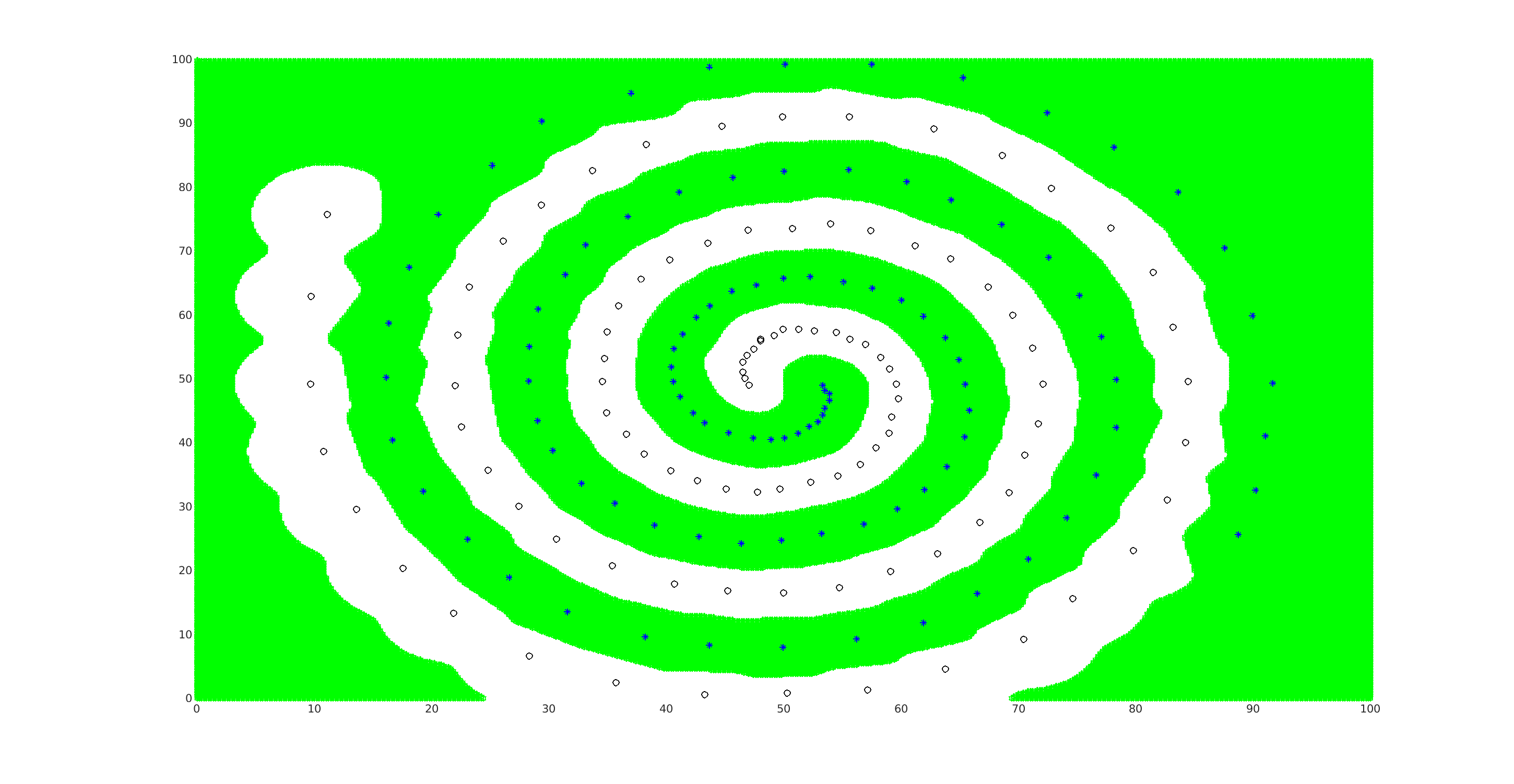} \\
\end{array}$$
\vskip-0.2cm
\caption{Spiral case}
\label{spiral}
\end{figure}

Three 2D tests with increasing difficulties are performed;  connected two sets, non-connected sets and finally a spiral case. Fig.\ref{connected}, \ref{non-connected},\ref{spiral} show that the new method separated successfully the data in all cases. The same value of $\sigma=1.2$ is used for the three tests.

For the high dimension case, two-classes tests are selected: 
Leukemia dataset on molecular classification of cancer \cite{Leu99}: training set 38 (C1=27, C2=11), testing set:34, features:7129. And svmguide1 dataset on astroparticle application \cite{svmguide103} set 3089 (C1=2000, C2=2000), testing set:4000, features:4

Table1 shows the performance comparison of the new method to optimized SVM. As we can see the method is performing as good the SVM with the following significant advantages:
\begin{enumerate}
\item The new algorithm is easier to implement, robust and faster since no optimization process is required that can slow down the SVM algrithm in high dimension and can fail even if the cost function is quadratic because of rouding-of errors.
\item The complexity for the SVM increase quadratically with the number of classes $n$ since we need  $n + \frac{n(n-1)}{2}$ hyper-plans to calculate, while for the new method the complexity is linear, $n$ distances need to be calculated for the decision.
\item Since the subspaces are independent calculated, the new method is highly parallelisable.
\item For dynamic classification, if a new class is created all hyperplanes in the output space have to be recalculated for the SVM, while for the new method the only thing that needs to be calculate is the new POD subspace.
\end{enumerate}

\begin{tabular}{ |p{3cm}|p{3cm}|p{3cm}|p{3cm}|  }
 \hline
 \multicolumn{4}{|c|}{Table1:The new method and SVM comparison} \\
 \hline
 \multicolumn{1}{|c|}{Test name/ Classes}
& \multicolumn{1}{|c|}{Test size}
& \multicolumn{1}{|c|}{SVM}
&\multicolumn{1}{|c|}{New Method} \\
 \hline
 Leukemia/2 & $Train. \#38  \newline Test. \# 34$ & $82.35 \% $ & $88.23 \% $ \\
 \hline
 svmguide1/2 & $Train. \#3089  \newline Test. \# 4000$ & $96.8 \% $ & $96.7 \% $ \\
 \hline
\end{tabular}

\section{Conclusions}
The paper proposed an alternative to SVM method based on a minimum distance to optimal subspaces containing the mapped attribute classes. The subspaces are obtained using POD method. The proposed method has the advantages of being robust, easy to implement, low complexity, designed for multi-classes, suitable for dynamic classifications and easy and efficient management of imbalanced classes.

\section*{Acknowledgement}
This research is supported by Alfaisal University grant IRG
with reference IRG20411.

\end{document}